\documentclass[conference,review]{IEEEtran}
\IEEEoverridecommandlockouts
\usepackage{cite}
\usepackage{amsmath,amssymb,amsfonts}
\usepackage{algorithmic}
\usepackage{graphicx}
\usepackage{textcomp}
\usepackage{xcolor}
\def\BibTeX{{\rm B\kern-.05em{\sc i\kern-.025em b}\kern-.08em
    T\kern-.1667em\lower.7ex\hbox{E}\kern-.125emX}}
    
\usepackage{multirow}
\usepackage{amsmath}
\usepackage{comment}
\usepackage{xspace}
\usepackage{graphicx}
\usepackage{subfigure}
\usepackage{epstopdf}
\usepackage{booktabs}
\usepackage{url}

\usepackage{stfloats}

\usepackage[section]{placeins}

\newcommand{\etal}{et al.\@\xspace}

\begin{document}

\title{Towards Generating Real-World Time Series Data
}

\makeatletter
\newcommand{\linebreakand}{%
  \end{@IEEEauthorhalign}
  \hfill\mbox{}\par
  \mbox{}\hfill\begin{@IEEEauthorhalign}
}
\makeatother

\author{\IEEEauthorblockN{1\textsuperscript{st} Hengzhi Pei}\thanks{The work was conducted when Hengzhi Pei was doing internship at Microsoft Research Asia. The corresponding authors are Kan Ren and Dongsheng Li.}
\IEEEauthorblockA{
 \textit{University of Illinois at Urbana-Champaign}\\
Urbana, US \\
hpei4@illinois.edu}\\
\and
\IEEEauthorblockN{2\textsuperscript{nd} Kan Ren}
\IEEEauthorblockA{
\textit{Microsoft Research Asia}\\
Shanghai, China \\
kan.ren@microsoft.com}
\and
\IEEEauthorblockN{3\textsuperscript{rd} Yuqing Yang}
\IEEEauthorblockA{
\textit{Microsoft Research Asia}\\
Shanghai, China \\
yuqing.yang@microsoft.com}
\linebreakand
\IEEEauthorblockN{4\textsuperscript{th} Chang Liu}
\IEEEauthorblockA{
\textit{Microsoft Research Asia}\\
Beijing, China \\
chang.liu@microsoft.com}
\and
\IEEEauthorblockN{5\textsuperscript{th} Tao Qin}
\IEEEauthorblockA{
 \textit{Microsoft Research Asia}\\
Beijing, China \\
taoqin@microsoft.com}
\and
\IEEEauthorblockN{6\textsuperscript{th} Dongsheng Li}
\IEEEauthorblockA{
 \textit{Microsoft Research Asia}\\
Shanghai, China \\
dongsheng.li@microsoft.com}
}

\maketitle

\begin{abstract}
Time series data generation has drawn increasing attention in recent years. Several generative adversarial network (GAN) based methods have been proposed to tackle the problem usually with the assumption that the targeted time series data are well-formatted and complete. However, real-world time series (RTS) data are far away from this utopia, e.g., long sequences with variable lengths and informative missing data raise intractable challenges for designing powerful generation algorithms. In this paper, we propose a novel generative framework for RTS data --- \textit{RTSGAN} to tackle the aforementioned challenges. \textit{RTSGAN} first learns an encoder-decoder module which provides a mapping between a time series instance and a fixed-dimension latent vector and then learns a generation module to generate vectors in the same latent space. By combining the generator and the decoder, \textit{RTSGAN} is able to generate RTS which respect the original feature distributions and the temporal dynamics. 
To generate time series with missing values, we further equip \textit{RTSGAN} with an observation embedding layer and a decide-and-generate decoder to better utilize the informative missing patterns. 
Experiments on the four RTS datasets show that the proposed framework outperforms the previous generation methods in terms of synthetic data utility for downstream classification and prediction tasks. 
Our code is available at \url{https://seqml.github.io/rtsgan}.
\end{abstract}

\begin{IEEEkeywords}
Time series, data generation, missing values
\end{IEEEkeywords}

\section{Introduction}
Time series data recorded by sensors become ubiquitous in many areas including healthcare, agriculture, manufacturing and many others~\cite{madakam2015internet}. However, many of these data are highly confidential, e.g., patient records, which may cause privacy or accessibility concerns when using them~\cite{aggarwal2008privacy,Park2018,choi2017generating,yoon2019time,esteban2017real}. Generating synthetic data for applications, e.g., follow-up machine learning tasks has recently become one of the promising solutions~\cite{esteban2017real,choi2017generating}. 
Although lack of theoretical guarantee, recent works proved that the generated data can be resilient to membership inference attack~\cite{Park2018}, patient re-identification~\cite{choi2017generating}, etc. More importantly, the generated data are often of higher utility than anonymization/perturbation methods~\cite{choi2017generating}.

Generating realistic time series data is a challenging problem~\cite{xu2020cot}. 
A good generative model should not only capture the multidimensional distribution at each time point but also the temporal dynamics across time. 
Further, synthetic time series should reflect corresponding global features, e.g., static features like age and the labels of our interests such as mortality in clinical data. 
Recently, due to the success of Generative Adversarial Networks (GANs)~\cite{goodfellow2014generative} and its variants~\cite{arjovsky2017wasserstein, gulrajani2017improved}, it is natural to extend the GAN framework for time series data generation by applying recurrent neural networks (RNNs) as the generator and the discriminator. Following this paradigm, several works were proposed to solve the generation problems on time series data~\cite{mogren2016c,esteban2017real,yoon2019time, lin2020using, xu2020cot}. 
However, these works are usually targeted at generation for simple, well-formatted time series data, thus may not be applicable to real-world time series data generation as shown in our empirical studies. 

\begin{figure}[tb!]
  \centering
  \includegraphics[width=1\linewidth]{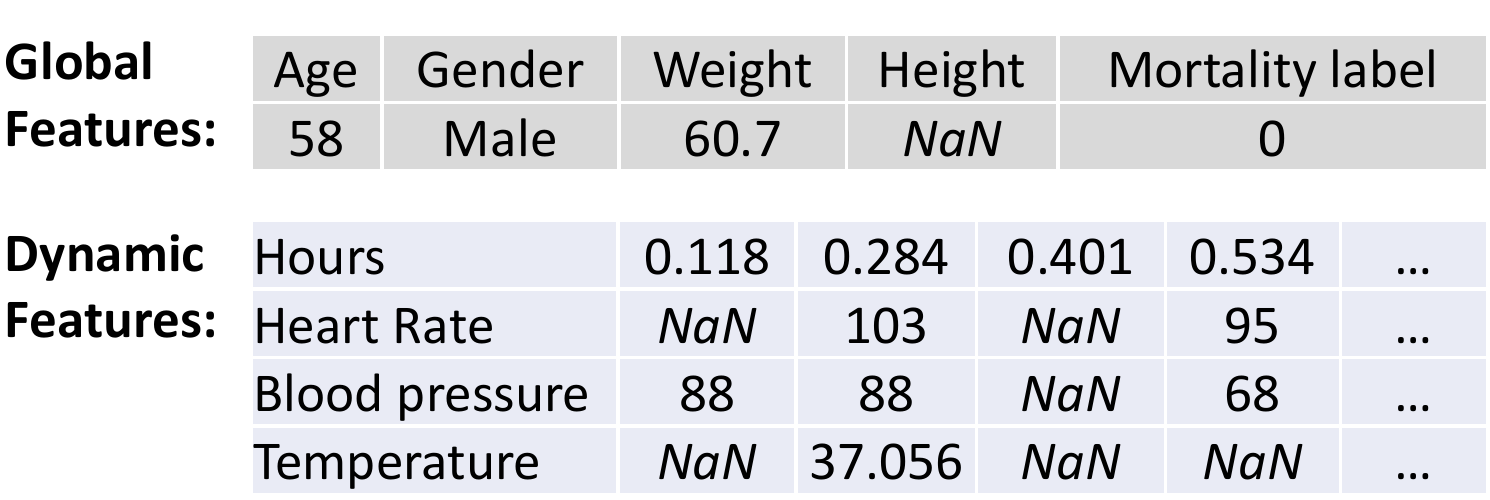}
  \caption{An example for real-world time series (\textit{NaN} denotes missing value).
  }
  \label{fig:example}
\end{figure}

The imperfect real-world time series data raise new challenges to the data generation algorithms. 1) Long sequences with variable lengths. The lengths of real-world time series data can be long, variable and sometimes critical, e.g., in survival analysis~\cite{ranganath2016deep,ren2019deep}.  However, many existing methods~\cite{esteban2017real, yoon2019time, xu2020cot} were only evaluated on short, fixed-length time series, leaving their performance on long, variable-length time series unexplored. 2) Missing values. Missing values are very common in real-world time series data. An example in clinical data for mortality prediction is shown in Figure \ref{fig:example}. These missing values can be informative~\cite{rubin1976inference}. For instance, missing values in clinical data can reflect the patient's situation and the physician's decision~\cite{sterne2009multiple}. And more and more works are focusing on exploiting the missing patterns to improve prediction performance~\cite{che2018recurrent, cao2018brits, luo2018multivariate, luo2019e2gan, tang2020joint, srivastava2020forecasting}. However, to the best of our knowledge, none of the existing works has considered generating time series with informative missing values.

In this paper, we propose the Real-world Time Series Generative Adversarial Network (\textit{RTSGAN}), a novel generative framework to tackle the aforementioned challenges. \textit{RTSGAN} consists of two key components: 1) an encoder-decoder module, which learns to encode each time series instance into a fixed-dimension latent vector and reconstruct the whole time series from the latent vector via an autoencoder; and 2) a generation module, in which a Wasserstein GAN (WGAN)~\cite{gulrajani2017improved} is trained to generate vectors in the same latent space of the above autoencoder. By using the generator and the decoder, \textit{RTSGAN} is able to generate real-world time series data which respect the original feature distributions and the temporal dynamics. To better address the informative missing issue, we extend \textit{RTSGAN} to \textit{RTSGAN-M}: an observation embedding is proposed to enrich the information at each time step, and a novel decide-and-generate decoder is also proposed which first decides the time and missing patterns of the next step and then generates the corresponding feature values based on both local and global dependencies. 
Empirical studies on four real-world time series datasets show that synthetic data generated by the proposed framework are not only more ``realistic-looking'', but also with higher utility for downstream machine learning tasks in the ``train on synthetic, test on real (TSTR)'' setting~\cite{esteban2017real}.

Our main contributions are summarized as follows:
\begin{itemize}
    \item We propose a novel time series data generation framework named \textit{RTSGAN} to tackle the challenges raised by real-world time series data. 
    \item To the best of our knowledge, this is the first work to investigate the problem of generating time series with missing values, in which an observation embedding and a novel decide-and-generate decoder in \textit{RTSGAN-M} are proposed to achieve better generation performance.
    \item Detailed experiments are conducted on four real-world datasets including both complete, fixed-length time series and incomplete, variable-length time series, in which \textit{RTSGAN} outperforms the state-of-the-art methods in terms of synthetic data utility in downstream classification and prediction tasks.
\end{itemize}

\section{Related Work}

Our work falls in the realm of time series data generation. Mogren~\cite{mogren2016c} first proposed the C-RNN-GAN method, using RNNs for both the generator and the discriminator to generate sequential data from random vector sequences. Using a similar architecture, RCGAN~\cite{esteban2017real} was then proposed to generate real-valued, labeled, medical data. However, traditional framework and loss function of GAN are not sufficient for multivariate time series generation because we need to capture not only the multidimensional feature distribution at each time point but also the temporal dependencies. Lin \cite{lin2020using} also identified these key challenges and designed a custom workflow called DoppelGANger. COT-GAN~\cite{xu2020cot} introduced a new adversarial objective based on optimal transport theories. But this method is unable to handle time series with variable lengths.

Autoencoders (AE) have also been exploited along with GANs,
which has shown success in computer vision \cite{makhzani2015adversarial} and natural language processing~\cite{zhao2018adversarially, haidar-etal-2019-latent}. For time series generation, TimeGAN \cite{yoon2019time} generates data from a learned embedding space which is optimized by binary adversarial feedback and stepwise supervised loss. 
The dimension of their embedding space is proportional to the sequence length because it still generates a sequence from a sequence of random vectors.
By contrast, we aim to generate a sequence from a latent space whose dimension is invariant of sequence length. In this respect, the proposed method is similar to RNN-based variational autoencoders (VAE) \cite{fabius2014variational, bowman2015generating}. However, the assumption that the latent distribution is Gaussian is not always realistic and restricts the generation performance because real-world data may follow a much more complex distribution.

Since we aim to generate time series with missing values, this work is also related to missing data analysis. 
Che~\etal~\cite{che2018recurrent} first introduced trainable decays to utilize missing patterns by incorporating masking and time intervals into a deep model and achieved better classification results. Efforts were later devoted to forecasting~\cite{tang2020joint, srivastava2020forecasting} and imputation~\cite{cao2018brits, luo2018multivariate, luo2019e2gan} for further exploiting informative missing values. Despite their success, these tasks are different from time series generation. Imputation aims to impute missing values for downstream analysis, and forecasting predicts feature values according to the past observations. Our goal is to generate time series with missing values which is more in line with real data. 

\section{Problem Formulation}
Generally, each instance in the time series data mainly consists of two types of features: dynamic features (which change over time, e.g., heart rate of one patient) and global features (which include static features, e.g., age, and the global properties of observed sequences, e.g., the label of our interests). 
Each feature in both dynamic features and global features can be continuous or categorical. We denote one instance in a time series training set $\mathcal{D}$ as $(\mathbf{X}, \mathbf{y})$. Here, $\mathbf{X} = (\mathbf{x}_1,...,\mathbf{x}_l)^\top \in \mathbb{R}^{l \times d_{x}}$ represents a $d_{x}$-dimensional multivariate time series which contains $l$ observations. $\mathbf{y} \in \mathbb{R}^{d_{y}}$ represents the global features of the time series.

In reality, time series data can be incomplete. In other words, missing values in time series data can commonly occur in both dynamic and global features as shown in Figure \ref{fig:example}. To formulate this problem, we denote each instance as $(\mathbf{X}, \mathbf{y}, \mathbf{M}^{(x)}, \mathbf{m}^{(y)})$. 
The mask matrix $\mathbf{M}^{(x)} \in \mathbb{R}^{l \times K}$ is introduced to represent the missing values for $K$ dynamic features, where $M^{(x)}_{i,j} = 1$ if $j$-th feature at the $i$-th observation is observed and 0 otherwise.
Similarly, $\mathbf{m}^{(y)}$ represents the missing values in the global features. 

The goal of this work is to use the training set $\mathcal{D}$ to learn data distribution and generate a synthetic dataset $\hat{\mathcal{D}}$ which is realistic-looking and has high utility. Downstream machine learning tasks, e.g., classification and sequence prediction, can be performed on the synthetic dataset $\hat{\mathcal{D}}$ and the resulting downstream models can have similar performance compared to those trained on $\mathcal{D}$.

\section{The Proposed RTSGAN Method}
\begin{figure}[tb!]
  \centering
  \includegraphics[width=1\linewidth]{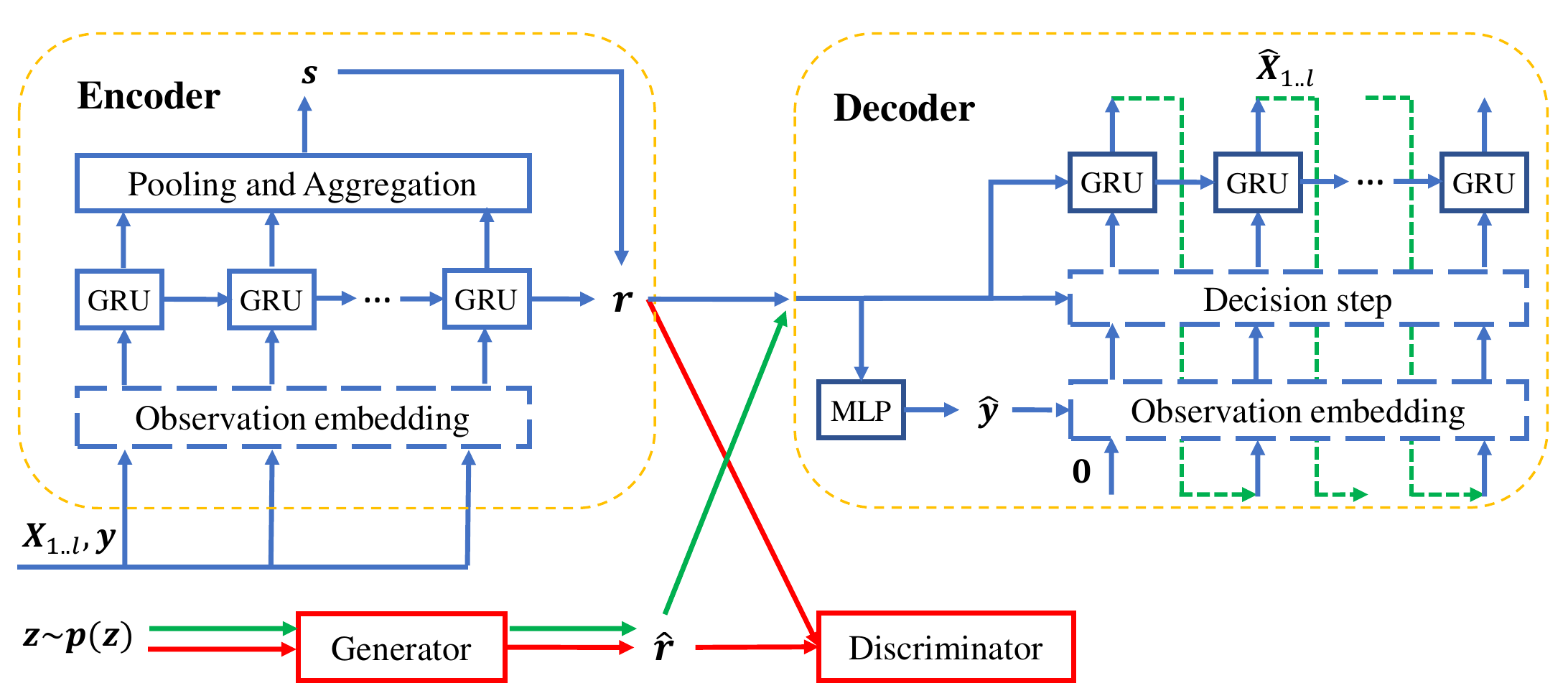}
  \caption{Architecture of the proposed \textit{RTSGAN}. The blue lines indicate the training phase of AE and the dashed blue lines indicate targeted designs of \textit{RTSGAN-M} for incomplete time series generation. The red lines indicate the training phase of WGAN. Once training is finished, the generation process flows through the green lines: first generate a latent representation $\hat{r}$ via the generator and then synthesize time series autoregressively via the decoder. 
  }
  \label{fig:RTSGAN}
\end{figure}

In this section, we present the real-world time series data generation framework --- \textit{RTSGAN}. As described in Figure~\ref{fig:RTSGAN}, \textit{RTSGAN} consists of two key components: 
\begin{itemize}
    \item {\bf Encoder-decoder module}. We first train an autoencoder to encode time series into a fixed-dimension latent space whose dimension is invariant of sequence length and then reconstruct inputs from this latent space; 
    \item {\bf Generation module}. After the training of encoder-decoder module is finished, we adopt the WGAN framework for generative modeling of the latent space so that the generator can output synthetic latent representations in the above latent space.
\end{itemize}
To generate time series, we just need to feed synthetic latent vectors from the generator to the decoder. 

Previous GAN-based methods usually requires discriminating a \textit{sequence} of vectors in either the feature space \cite{esteban2017real, lin2020using,xu2020cot} or the latent space \cite{yoon2019time}. By contrast, our framework only requires discriminating a \textit{single} latent vector. When the time series data become complicated, e.g., with variable lengths and missing values, it would be easier for \textit{RTSGAN} to capture the original data distributions. 

Next, we will describe the details of \textit{RTSGAN} on both complete time series and incomplete time series.

\subsection{RTSGAN on Complete Time Series}
We first present \textit{RTSGAN} for generating complete time series without missing values. 

\subsubsection{The Encoder-Decoder Module}

This module consists of an encoder to encode the input sequence into a latent vector and a decoder to reconstruct the input sequence from the latent vector. Before feeding into the autoencoder, all features are transformed into $[0, 1]$, using min-max scaling for continuous features and one-hot encoding for categorical features. 

{\bf Encoder}. Unlike the encoder in TimeGAN \cite{yoon2019time} which encodes each time series into a sequence of latent vectors, the encoder in our design aims to encode each time series into a compact representation whose dimension is invariant of sequence length. We first concatenate the global features $\mathbf{y}$ on the dynamic features $\mathbf{x}_i$ at each step as $\mathbf{e}_i = [\mathbf{x}_i, \mathbf{y}]$, and feed it into an $N$-layer Gated Recurrent Unit (GRU) \cite{cho2014learning} with hidden dimension $d_\mathrm{AE}$ to get the hidden states $\mathbf{h}_{i}^{n}$ of each step $i$ for each GRU layer $n$ as follows:
\begin{equation}
  \mathbf{h}_{i}^{n} = \mathrm{GRU}(\mathbf{e}_{i}), \ i \in [1,\ l], n\in [1,\ N].
\end{equation}
To better capture the temporal dynamics and global properties of time series, 
we further apply pooling operations on the hidden states from the last layer of GRUs $\mathbf{h}_{i}^{N}$ to enrich the representation as follows:
\begin{equation}
  \mathbf{s} = \mathrm{FC}([\mathrm{AvgPool}(\mathbf{h}_{i}^{N}),\ \mathrm{MaxPool}(\mathbf{h}_{i}^{N}),\ \mathbf{h}_{l}^{N}]) 
\end{equation}
Here, $\mathrm{FC}$ denotes a fully connected layer which aggregates the pooling results into space $\mathbb{R}^{d_\mathrm{AE}}$, and $\mathrm{LeakyReLU}$ \cite{maas2013rectifier} is used as activation function. Then, we concatenate the global information $\mathbf{s}$ and the last hidden state together to get a latent representation $\mathbf{r} \in \mathbb{R}^{(N+1)d_\mathrm{AE}}$ for a time series as follows:
\begin{equation}
  \mathbf{r} = [\mathbf{s},\ \{\mathbf{h}_{l}^{n}\}_{n=1}^{N}].
\end{equation}

{\bf Decoder}. The decoder aims to reconstruct the whole time series from the latent representation $\mathbf{r}$. It includes two steps: 1) first reconstruct the global features $\hat{\mathbf{y}}$ via a fully connected layer and 2) then reconstruct the dynamic features $\mathbf{h}_{l}^{n}$ via a GRU. The global features $\hat{\mathbf{y}}$ is reconstructed as follows: 
\begin{equation}
  \hat{\mathbf{y}} = \mathrm{Act}(\mathbf{W}_{y} \mathbf{s} + \mathbf{b}_y). \\
\end{equation}
In the $\mathrm{Act}$ function, we apply softmax for categorical features and sigmoid for continuous features.

Next, we reconstruct the dynamic features. The decoder for the dynamic features is another $N$-layer GRU with hidden dimension $d_\mathrm{AE}$ which takes $\mathbf{h}_{l}^{n}$ as the initial hidden state $\hat{\mathbf{h}}_{0}^{n}$. The reconstruction process is autoregressive aiming to model each of $p(\mathbf{x}_i|\mathbf{x}_{1..i-1}, \mathbf{y})$ as follows:
\begin{equation}
\begin{aligned}
    & \hat{\mathbf{e}}_{i} = [\mathbf{x}_{i-1},\ \mathbf{s}], \ 
  \hat{\mathbf{h}}_{i}^{n} = \mathrm{GRU}(\hat{\mathbf{e}}_{i}, \hat{\mathbf{h}}_{i-1}^{n}), \ n\in [1,\ N]. \\
  & \hat{\mathbf{x}}_{i} = \mathrm{Act}(\mathbf{W}_{x} \hat{h}_{i}^{N} + \mathbf{b}_x). \\
\end{aligned}
\end{equation}
The initial input at the beginning of the autoregressive process is $\hat{\mathbf{e}}_{1} = [\mathbf{0}, \mathbf{s}]$. To deal with time series with variable lengths, we include sequence length $l$ as one of the global features. So after global features are reconstructed, we can control the dynamic feature reconstruction exactly by $\hat{l}$.

For the training of the autoregressive recurrent networks, we can use teacher-forcing \cite{graves2013generating, sutskever2011generating} which always uses ground-truth data $\mathbf{x}_{i-1}$ as the next-step input, or sampling \cite{bengio2015scheduled} on previous prediction $\hat{\mathbf{x}}_{i-1}$ and ground truth $\mathbf{x}_{i-1}$. The overall loss function is a linear combination of reconstruction loss for global features and dynamics features:
\begin{equation}
\label{eqn:loss}
  L_\mathrm{re} = \frac{d_y}{d_x+d_y}L_{y}(\hat{\mathbf{y}}, \mathbf{y}) + \frac{d_x}{d_x+d_y}L_{x}(\hat{\mathbf{X}}, \mathbf{X}).
\end{equation}
Cross-entropy (CE) loss and mean squared error (MSE) loss are used for categorical features and continuous features respectively.

\subsubsection{The Generation Module}

As shown above, both global features and dynamic features are encoded into the same latent space, so that $\mathbf{r}$ naturally contains various relations inside time series and the autoregressive decoder itself maintains the temporal dynamics of time series. Therefore, we can synthesize a representation in the latent space and decode it autoregressively to generate the whole time series instead of producing synthetic outputs directly from the feature space. 

Since the dimension of the latent space is invariant of sequence length $l$, it is much easier for the generation module to synthesize latent representations. Here, we employ the improved version of WGAN~\cite{gulrajani2017improved}. The generator in WGAN aims to minimize the 1-Wasserstein distance $W(P_{r}, P_{g})$ between real data distribution and synthetic data distribution, with the help of an iteratively trained 1-Lipschitz discriminator. The optimization objective of the WGAN is defined as follows:
\begin{equation}
  \min_{G} \max_{D} E_{\mathbf{r} \sim \mathrm{encoder}(\mathbf{X}, \mathbf{y})}[D(\mathbf{r})] - E_{\mathbf{z} \sim p(\mathbf{z})}[D(G(\mathbf{z}))].
\end{equation}
Here, $G$ and $D$ denote the generator and the 1-Lipschitz discriminator respectively. In practice, we use a multi-layer perception (MLP) with layer normalization \cite{ba2016layer} for $G$, and three fully connected layers for $D$. $\mathrm{LeakyReLU}$ \cite{maas2013rectifier} is used as activation function for both $G$ and $D$. After the training of WGAN, we can generate time series data as follows: 
\begin{equation}
\hat{\mathbf{X}}, \hat{\mathbf{y}} = \mathrm{decoder}(G(\mathbf{z})),~ \mathbf{z} \sim p( \mathbf{z} ).
\end{equation}
Some previous AE-GAN-based generation methods~\cite{choi2017generating, spinks2018generating, haidar-etal-2019-latent} require fine-tuning the parameters of the decoder during the process of GAN training because they need to further discriminate real data and synthetic data on the feature space. However, we find that the proposed method can achieve good generation performance without fine-tuning when only discriminating data on the latent space. So we just train the encoder-decoder module and the generation module separately to simplify the training process.

\subsection{RTSGAN on Incomplete Time Series}
We now extend \textit{RTSGAN} to generate incomplete time series, in other words, to generate time series with missing values. 
The main idea is to generate both missing vector and feature vector at each time step, then mask the corresponding feature values according to the generated missing vector. A simple way to achieve this is to treat missing information of each feature as an additional binary feature and generate it as complete time series. 
However, the high rate of missing values in the real-world time series data may result in a catastrophic collapse in traditional GAN training since directly discriminating incomplete time series may not provide informative signals for the generator, which has been empirically shown in our experiments.
It also poses a challenge for the generative models to find the underlying correlations between the missing values and the observed values.

To this end, we propose a variant of \textit{RTSGAN} named \textit{RTSGAN-M}, in which the same AE-GAN framework is adopted but with new designs for better generation performance on time series data with missing values. More specifically, we propose two techniques in the encoder-decoder module: 1) observation embedding, which can enrich information at each observation; and 2) decide-and-generate decoder, which first decides the time and missing patterns of the next observation and then generates the corresponding feature values based on both local and global dependencies. Note that the generation module is unchanged in \textit{RTSGAN-M}.

\subsubsection{Observation embedding}

As mentioned above, high missing rates (e.g., $\sim$80\% missing in the PhysioNet dataset~\cite{silva2012predicting}) can make it difficult for the generative models to fully capture useful information from time series. Therefore, before feeding time series into the encoder-decoder module, we add an observation embedding layer to enrich the representations of time series at each step.

First, it is natural to consider the last valid observations at each step because some features can be stable which will not be measured frequently. Second, the time point $t_i$ of each observation should be emphasized especially in irregular sampled time series, because the time points of the observations are often related to missing rates ~\cite{luo2018multivariate} and time intervals often reflect the influence of the previous observations. Despite the capability to capture the order of sequences, the RNNs may not be sensitive enough to capture information from time intervals~\cite{shukla2020multi}. So similar to the positional encoding used in Transformer~\cite{vaswani2017attention}, we need an embedding layer for the time points to fully exploit temporal information.

Therefore, we build an observation embedding by using the feature values, missing patterns and time point as follows:
\begin{equation}
  \mathbf{e}_{i} = \mathbf{W} [\mathbf{x}_i, \mathbf{pre}_i, \mathbf{M}^{(x)}_i, \mathbf{y}, \mathbf{m}^{(y)}] + \phi(t_i),
\end{equation}
where $\mathbf{pre}_i$ denotes the last observations of the dynamic features before the $i$-th observation. The $\phi(t_i)$ therein is a learnable time representation which is proposed by \cite{kazemi2019time2vec}:
\begin{equation}
\phi(t)[j] =
\begin{cases}
w_0 t+b_0 & j==0, \\
\sin (w_j t+b_j)& \text{otherwise}.
\end{cases}
\end{equation}
By doing so, it will be easier for the encoder-decoder module to capture the relations among time points, missing values and observed values. The parameters of the overall observation embedding are shared between the encoder and the decoder.

\subsubsection{Decide-and-generate Decoder}

In many real-world applications, informative missing values can be related to sampling decisions~\cite{sterne2009multiple}, e.g., a physician may decide which dynamic features should be measured next time according to a patient's situation. Therefore, it is more reasonable to model $p(\mathbf{x}_i, \mathbf{M}^{(x)}_i|\mathrm{info}_{i-1})$ by the following conditional distribution:
\begin{equation}
  p(\mathbf{x}_i, \mathbf{M}^{(x)}_i|\mathrm{info}_{i-1}) = p(\mathbf{x}_i | \mathbf{M}^{(x)}_i, \mathrm{info}_{i-1}) p(\mathbf{M}^{(x)}_i|\mathrm{info}_{i-1}). \\
\end{equation}
where $\mathrm{info}_i := \{\mathbf{x}_{1..i}, \mathbf{M}^{(x)}_{1..i}, \mathbf{y}, \mathbf{m}^{(y)}\}$ denotes all the information until the $i$-th observation.   $p(\mathbf{x}_i|\mathbf{M}^{(x)}_i, \mathrm{info}_{i-1})$ models the dynamic feature distribution conditioned on the other information and $p(\mathbf{M}^{(x)}_i|\mathrm{info}_{i-1})$ models the missing data distribution conditioned on the previous information.

Following the above idea, we split the dynamic reconstruction into two steps: decide and generate. The decision step consists of $N_\mathrm{dec}$-layer GRU and generates the time point and masks of the next observation as follows:
\begin{equation}
\begin{aligned}
  & \hat{\mathbf{e}}_{i} = W [\mathbf{x}_{i-1}, \mathbf{pre}_{i-1}, \mathbf{M}^{(x)}_{i-1}, \mathbf{y}, \mathbf{m}^{(y)}] + \phi({t}_{i-1}). \\
  & \hat{\mathbf{h}}_{i}^{n} = \mathrm{GRU}(\hat{\mathbf{e}}_{i}, \hat{\mathbf{h}}_{i-1}^{n}), \  n \in [1,\ N_\mathrm{dec}]. \\
  & \hat{t}_i = \sigma(\mathbf{W}_t \hat{\mathbf{h}}_{i}^{N_\mathrm{dec}} + b_t) + t_{i-1}. \\
  & \hat{\mathbf{M}}^{(x)}_i = \sigma(\mathbf{W}_{m} \hat{\mathbf{h}}_{i}^{N_\mathrm{dec}} + \mathbf{b}_{m}). \\
\end{aligned}
\end{equation}
With certain thresholds, we can decide which feature values will be generated at the $i$-th observations according to $\hat{\mathbf{M}}^{(x)}_{i}$. After making a decision about $\hat{t}_i$ and $\hat{\mathbf{M}}^{(x)}_{i}$, we need to model $p(\mathbf{x}_i|\mathbf{M}^{(x)}_i, \mathrm{info}_{i-1})$ by taking a generation step. Here, we introduce the concept of time lag: time interval between two consecutive valid observations for a feature. Formally, we calculate the time lag matrix $\mathbf{\delta} \in \mathbb{R}^{l \times K}$ for $K$ dynamic features as follows:
\begin{equation}
\delta_{i, j} =
\begin{cases}
t_i - t_{i-1} & \mathbf{M}^{(x)}_{i-1,j}==1, \\
\delta_{i-1, j} + t_i - t_{i-1} & \mathbf{M}^{(x)}_{i-1,j}==0 \ \text{and } i>0, \  \\
0 & i==0. \\
\end{cases}
\end{equation}
The previous works~\cite{che2018recurrent, luo2018multivariate} have shown that changeable time lags are important and the influence of the past observations should be reduced if some features have been missing for a long time. Following this idea, we also introduce trainable decays as follows:
\begin{equation}
\begin{aligned}
  & \mathbf{\beta}_{i} = e^{-\max(\textbf{0}, W_{\beta}[\mathbf{\delta}_{i}, \  \mathbf{M}^{(x)}_i] + \mathbf{b}_{\beta})}. \\
& \mathbf{q}_i = \beta_{i} \odot \hat{\mathbf{h}}_{i}^{N_\mathrm{dec}}.
\end{aligned}
\end{equation}
$\mathbf{q}_i$ is the estimated local information at the next time point $\hat{t}_{i}$. We use it along with the global information $\mathbf{s}$ to generate the dynamic feature values with $(N - N_\mathrm{dec})$-layer GRU:
\begin{equation}
\begin{aligned}
  & \hat{\mathbf{h}}_{i}^{n} = \mathrm{GRU}([\mathbf{q}_i, \mathbf{s}], \hat{\mathbf{h}}_{i-1}^{n}), \  n \in [N_\mathrm{dec}+1,\ N]. \\
  & \hat{\mathbf{x}}_{i} = \mathrm{Act}(\mathbf{W}_{x} \hat{\mathbf{h}}_{i}^{N} + \mathbf{b}_x).
\end{aligned}
\end{equation}
Recall that $\mathbf{s}$ comes from the pooling operations on the whole time series and later is supervised by the global features, we can explore both local and global dependencies simultaneously for better feature reconstruction using $\mathbf{q}_i$ and $\mathbf{s}$.

\subsubsection{Loss Function with Missing Values Handling}

Here, we present the loss function for training the encoder-decoder module under the incomplete time series data setting. It consists of a feature reconstruction loss and a missing reconstruction loss. For feature reconstruction of incomplete time series, we only calculate the loss of valid observations. For missing reconstruction, we use binary cross-entropy (BCE) as the loss function. Since missing values of some features can frequently happen, we should try to maintain the observations for the dynamic features which are seldom observed in each time series. Thus, we set the rescaling weight $w_{i, j}$ for the dynamic feature $x_{i,j}$  according to the overall missing rate $\rho_{j}$ of the $j$-th dynamic feature as follows: 
\begin{equation}
\begin{aligned}
& w_{i, j} \propto (1 - \rho_j) \cdot (1 - \mathbf{M}^{(x)}_{i, j}) + \rho_j \cdot \mathbf{M}^{(x)}_{i, j}. \\
& L_{Mx} = \frac{1}{K \cdot l} \sum_{j=1}^{K} \sum_{i=1}^{l} w_{i,j} \cdot \mathrm{BCE}(\hat{\mathbf{M}}^{(x)}_{i,j}, \mathbf{M}^{(x)}_{i,j}). \\
\end{aligned}
\end{equation}
The new loss function for dynamic features are as follows:
\begin{equation}
  L_{x}^{'}= L_{x}(\hat{X}, X, \mathbf{M}^{(x)}) + L_{Mx}(\hat{\mathbf{M}}^{(x)}, \mathbf{M}^{(x)}, w).
\end{equation}
The loss for global features $L_{y}^{'}$ can be similarly decomposed into two components as above. Then, the overall loss function can be derived from Equation~\ref{eqn:loss} as
$L_\mathrm{re}^{'} = \frac{d_y}{d_x+d_y}L_{y}^{'}(\hat{\mathbf{y}}, \mathbf{y}) + \frac{d_x}{d_x+d_y}L_{x}^{'}(\hat{\mathbf{X}}, \mathbf{X}). $

\section{Experiments}
We evaluate \textit{RTSGAN} in two aspects: 1) realistic-looking, i.e., if the distributions of synthetic data are consistent with those of real data, and 2) high-utility, i.e., if the models trained on synthetic data can achieve decent performance on real data.   

\subsection{Complete, Fixed-Length Time Series}

\subsubsection{Datasets}

We first compare the performance of \textit{RTSGAN} against previous methods on two real-world complete, fixed-length time series datasets as follows.

\textbf{Stocks}.  We use the daily historical Google stocks data from 2004 to 2019\footnote{https://finance.yahoo.com/quote/GOOG/history?p=GOOG}. Each time step consists of 6 continuous-valued features: the volume and high, low, opening, closing, and adjusted closing prices. The sequences are aperiodic but features are highly correlated.

\textbf{Energy}.  We use the UCI Appliances energy prediction dataset~\cite{candanedo2017data} which contains regular measurements over time. Apart from timestamps, there are 28 correlated features at each time step, including temperature, humidity, energy usage and weather information from a nearby airport. 

\subsubsection{Settings}
Following the setting from the previous work \cite{yoon2019time}, the length of time series from these two datasets is set as 24. In total, 3,773 sequences are extracted from Stocks and 19,711 sequences are extracted from Energy. For a fair comparison, we use the same 3-layer GRU with hidden dimension 4 times the size of input features ($4\times d_x$) for the autoencoder of \textit{RTSGAN}. Since there is no global feature $y$ in these two datasets, we ignore that part in \textit{RTSGAN}. The detailed hyper-parameter setting can be found in the Appendix~\ref{appendix:complete}. 

\textbf{Compared Methods}.
We compare \textit{RTSGAN} with COT-GAN~\cite{xu2020cot}, TimeGAN~\cite{yoon2019time}, RCGAN~\cite{esteban2017real}, C-RNN-GAN~\cite{mogren2016c} for complete time series generation. Additionally, we consider the performance of WaveNet~\cite{oord2016wavenet} and its GAN counterpart WaveGAN \cite{donahue2018adversarial}. We follow the same metrics used in~\cite{yoon2019time} to measure the diversity, fidelity and utility of synthetic data from both quantitative and qualitative perspectives.

\textbf{Quantitative evaluation}. We adopt the following metrics to quantitatively evaluate the synthetic data.  \textbf{a) Discriminative Score}. It is a quantitative measure of similarity between real data and synthetic data. We train a 2-layer LSTM \cite{hochreiter1997long} to distinguish between time series from the real and synthetic datasets as a supervised learning task. We can then get classification accuracy on the held-out test set and the discriminative score is $|0.5 - \mathrm{accuracy}|$. Lower discriminative score indicates higher similarity between the real and synthetic datasets.
\textbf{b) Predictive Score}. To evaluate how well the synthetic data maintain the temporal dynamics, we perform next-step prediction task under the ``train on synthetic, test on real'' (TSTR) setting \cite{esteban2017real}. We train a 2-layer LSTM predictor and evaluate the model on the original dataset. Predictive score is calculated by the mean absolute error (MAE) and a lower score indicates better performance.

\textbf{Qualitative evaluation}. We apply t-SNE \cite{maaten2008visualizing} and PCA \cite{bryant1995principal} on both the real and synthetic datasets after flattening the temporal dimension. This gives us qualitative assessments of how close the distribution of the synthetic data is to that of the real data in 2D space.

\subsubsection{Results}
\begin{table}[]
\centering
\caption{Performance comparison on complete, fixed-length time series datasets. Boldface indicates the best discriminative score and predictive score (lower the better).}
\label{tab:complete}
\resizebox{\linewidth}{!}{\large
\begin{tabular}{ccccc}
\toprule
\multirow{2}{*}{Methods} & \multicolumn{2}{c}{Stocks}              & \multicolumn{2}{c}{Energy} \\
\cmidrule(lr){2-3} \cmidrule(lr){4-5}
~ & Discriminative     & Predictive         & Discriminative     & Predictive \\
\midrule
RCGAN                                       & 0.196±0.027          & 0.040±0.001          & 0.336±0.017          & 0.292±0.005          \\
C-RNN-GAN                                   & 0.399±0.028          & 0.038±0.000          & 0.499±0.001          & 0.483±0.005          \\
WaveNet                                     & 0.232±0.028          & 0.042±0.001          & 0.397±0.010          & 0.311±0.005          \\
WaveGAN                                     & 0.217±0.022          & 0.041±0.001          & 0.363±0.012          & 0.307±0.007          \\
TimeGAN                                     & 0.102±0.021          & 0.038±0.001          & 0.236±0.012          & 0.273±0.004          \\
COT-GAN                                      & 0.083±0.036          & 0.038±0.000          & 0.389±0.049          & 0.269±0.002          \\
RTSGAN (ours)                                  & \textbf{0.024±0.024} & \textbf{0.037±0.000} & \textbf{0.228±0.006} & \textbf{0.252±0.000} \\
\midrule
Original                                    &                    & 0.036±0.001          &                    & 0.250±0.003  \\       \bottomrule
\end{tabular}
}
\end{table}

\begin{figure*}[h]
  \centering
  \subfigure{\includegraphics[width=0.49\linewidth]{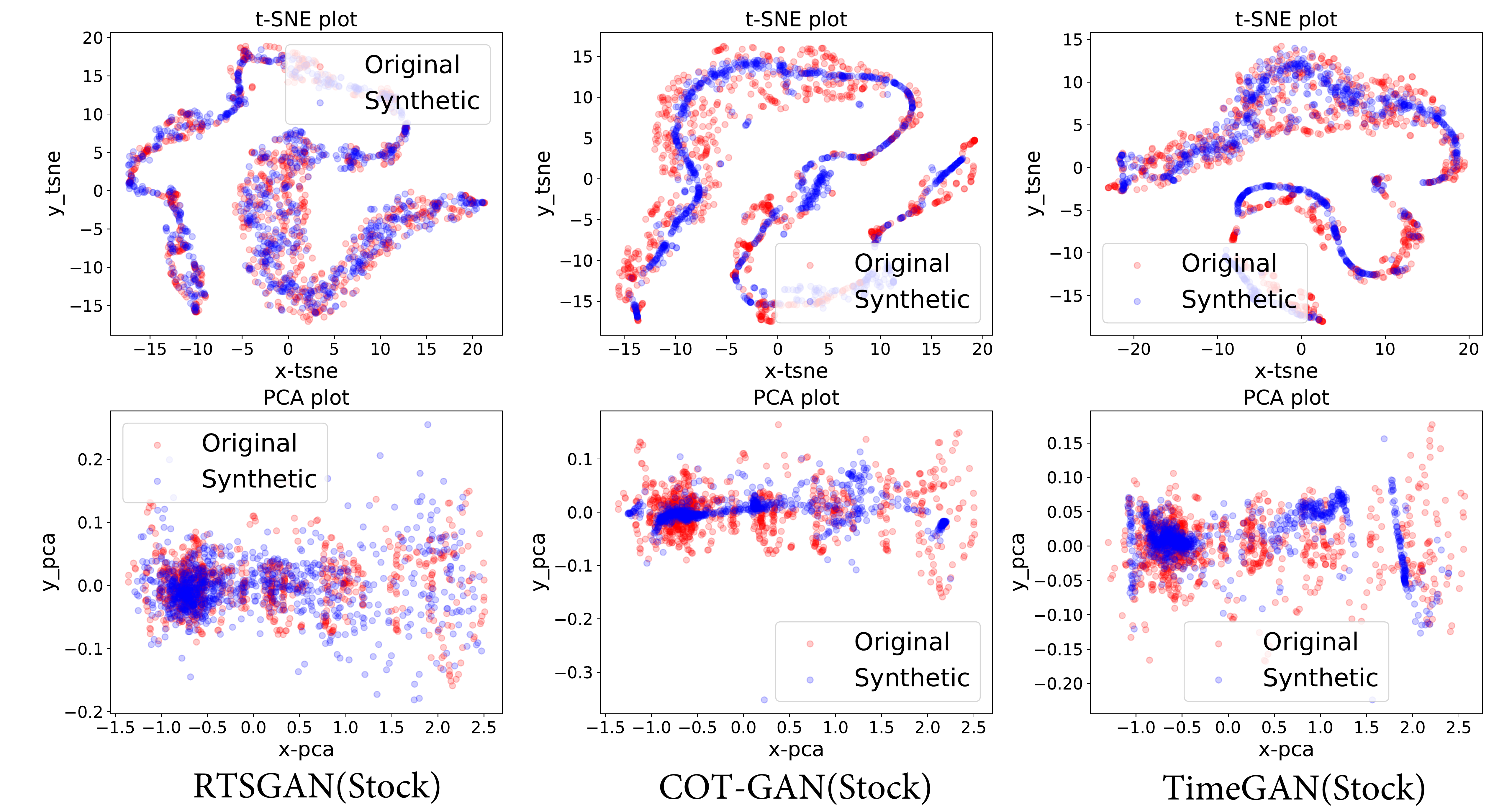}}
  \subfigure{\includegraphics[width=0.49\linewidth]{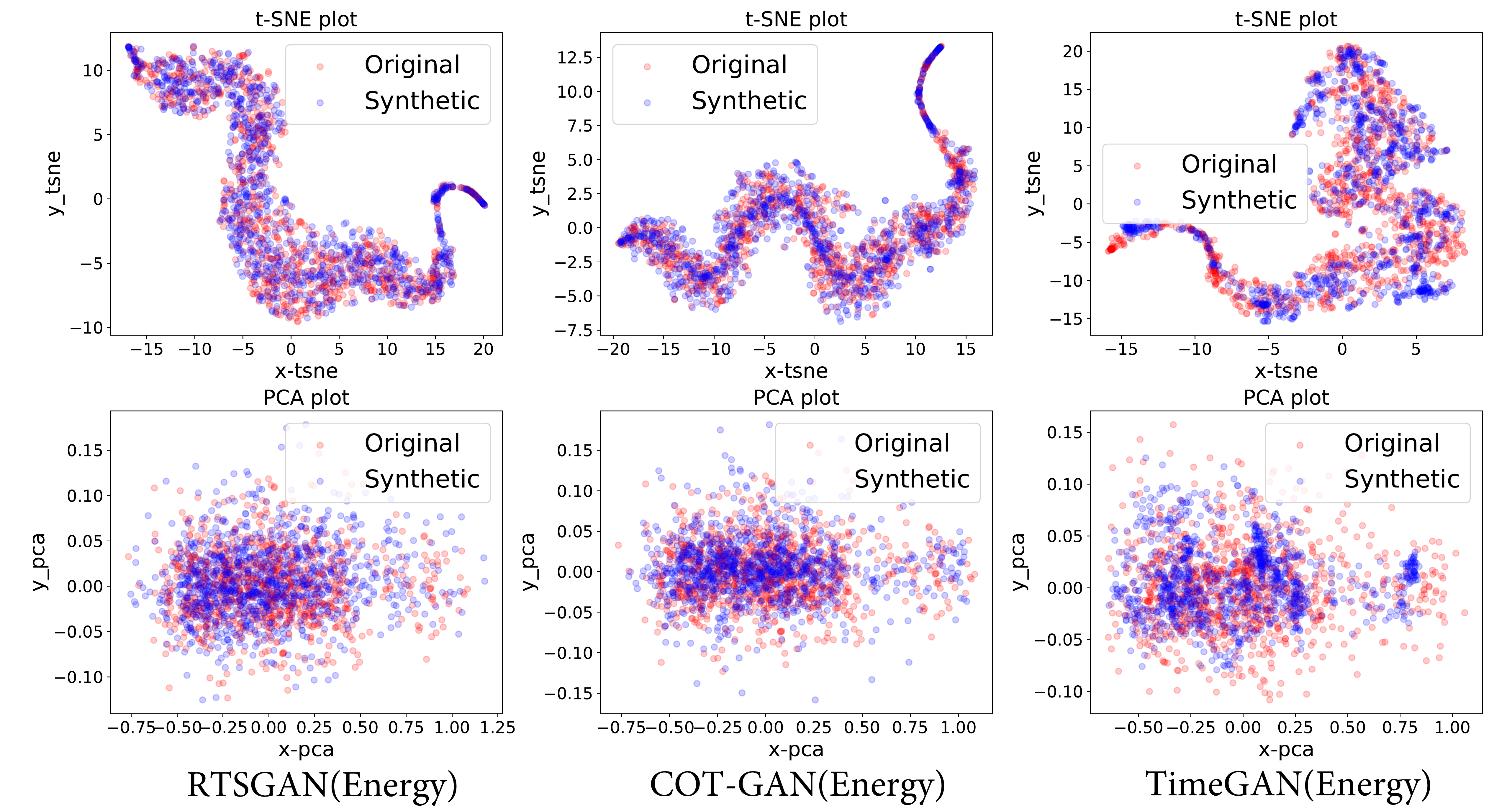}}
  \caption{t-SNE (top row) and PCA (bottom row) visualization results on Stocks (left 3 columns) and Energy (right 3 columns). Red and blue points denotes the original data and the synthetic data respectively. }\label{fig:visualization}
\end{figure*}

As shown in Table \ref{tab:complete}, \textit{RTSGAN} achieves the state-of-the-art results on both datasets in terms of both discriminative score and predictive score. Remarkably, the predictive scores of \textit{RTSGAN} are much better than previous methods and nearly in line with those of the original datasets. This demonstrates that the synthetic time series data generated by the proposed method has much higher utility for the sequence prediction task.

Figure~\ref{fig:visualization} shows the t-SNE and PCA visualization results on the Stocks and Energy datasets. We can find that the synthetic data generated by \textit{RTSGAN}  match the original distribution better than the previous state-of-the-art method. In addition, we can observe that our synthetic data is more diverse, which can help to explain why \textit{RTSGAN} achieves better discriminative scores (i.e., harder to discriminate) in Table \ref{tab:complete}.

\subsection{Incomplete, Variable-Length Time Series}

\begin{table*}[]
\caption{Statistics of the two real-world time series datasets with missing data.}
\label{tab:statistic}
\begin{tabular}{cccccccc}
\toprule
Dataset   & \# of train set & \# of test set & missing rate & \# of dynamic features & dynamic feature type & \# of static features & Avg.Len \\
\midrule
Physionet & 3600           & 400           & 80.67\%       & 36                    & continuous               & 5                    & 74      \\
MIMIC-III & 17737          & 3236          & 63.68\%       & 17                    & continuous, categorical                    & 0                    & 77   \\
\bottomrule
\end{tabular}
\end{table*}

\subsubsection{Datasets}

Next we evaluate the performance of \textit{RTSGAN} on variable-length time series with informative missing values using two real-world clinical datasets.

\textbf{PhysioNet Challenge 2012 dataset (PhysioNet).}~\cite{silva2012predicting} It is a public electronic dataset which provides 4000 multivariate clinical time series from the intensive care unit (ICU). Every ICU stay is a time series with 41 features which contains both static features and dynamic features. The task is mortality prediction, i.e., predicting whether a patient dies in the hospital. 

\textbf{MIMIC-III.} Harutyunyan~\etal~\cite{harutyunyan2019multitask} provided healthcare benchmarks using the data from Medical Information Mart for Intensive Care (MIMIC-III) database \cite{johnson2016mimic}. Here, we also use the data from the in-hospital mortality prediction task which contains the measurements of the first 48 hours of an ICU stay. We merge the categories with the same meaning for categorical features and set outlier ranges for continuous features to mitigate the data noises. 

More detailed statistics about the two datasets can be found in Table~\ref{tab:statistic}. We can see that both these two real-world datasets are with high missing rates and long average sequence lengths, which raises challenges for data generation.

\begin{table*}
\centering
\caption{Dataset utility comparison on the MIMIC-III dataset. Boldface indicates the best AUC performance.}
\label{tab:mimic3}
\begin{tabular}{ccccccccc}
\toprule
\multirow{2}{*}{Methods} & \multicolumn{3}{c}{min-max scaling} & \multicolumn{4}{c}{standard scaling}       & \multirow{2}{*}{Avg.} \\
\cmidrule(lr){2-4} \cmidrule(lr){5-8}
                         & LR       & zeroRNN   & lastRNN   & LR     & zeroRNN & lastRNN & discreteLSTM &                      \\
\midrule
TimeGAN & 0.6240 & 0.6171 & 0.6098 & 0.5583 & 0.5670 & 0.5767 & 0.5595 & 0.5875 \\
DoppelGANger & 0.7632 & 0.6377 & 0.6535 & 0.6553 & 0.6245 & 0.5751 & 0.5214 & 0.6330 \\
RTSGAN (ours) & 0.8009 & 0.7755 & 0.7682 & 0.6609 & 0.7450 & 0.7532 & 0.7940 & 0.7568               \\
RTSGAN-M (ours) & {\bf 0.8209} & {\bf 0.8108} & {\bf 0.8068} & {\bf 0.7769} & {\bf 0.7753} & {\bf 0.7893} & {\bf 0.8221} & {\bf 0.8003}
  \\
\midrule
Original                      & 0.8242   & 0.8424    & 0.8388    & 0.8509 & 0.8574  & 0.8509  & 0.8550      & 0.8457 \\     \bottomrule
\end{tabular}
\end{table*}

\begin{table}[]
\centering
  \caption{Dataset utility comparison on the PhysioNet dataset. Boldface indicates the best AUC performance.}
\label{tab:compare2012}
\resizebox{\linewidth}{!}{\large
\begin{tabular}{cccccc}
\toprule
\multirow{2}{*}{Methods} & \multicolumn{2}{c}{min-max scaling} & \multicolumn{2}{c}{standard scaling} & \multirow{2}{*}{Avg.} \\
\cmidrule(lr){2-3} \cmidrule(lr){4-5}
                        & zeroRNN            & lastRNN           & zeroRNN             & lastRNN            &                      \\
\midrule
TimeGAN  & 0.6575 & 0.6885 & 0.6493 & 0.6282 & 0.6558 \\ 
DoppelGANger  & 0.5794 & 0.6375 & 0.5266 & 0.5356 & 0.5698  \\
RTSGAN (ours)                 & 0.7866 & 0.8172 & 0.7371 & 0.6555 & 0.7491   \\
RTSGAN-M (ours)                & {\bf 0.8152} & {\bf 0.8242} & {\bf 0.8134} & {\bf 0.7857} & {\bf 0.8096}   \\
\midrule
Original                     & 0.8317          & 0.8414         & 0.8291   & 0.8353  & 0.8344    \\
\bottomrule
\end{tabular}
}
\end{table}

\subsubsection{Settings}
Since this is the first work to investigate time series generation with missing values, there is no baseline method to compare with. Thus, we modify existing methods to deal with this setting as follows: we treat missing patterns as additional binary features and concatenate them with the original time series. The missing values are imputed with zeros after data scaling. We also tried imputation with the last valid observations but it did not affect the overall generation performance. Then, we treat the pre-processed time series as complete time series for data generation. After generation, we transform the synthetic time series into its incomplete version according to the synthetic missing patterns. 

In this setting, we compare \textit{RTSGAN} against TimeGAN~\cite{yoon2019time} and DoppelGANger \cite{lin2020using} because they are designed to handle both static and dynamic features of time series, and they can also generate time series with flexible lengths. For \textit{RTSGAN} with the observation embedding and decide-and-generate decoder, we name it \textit{RTSGAN-M} to differentiate it from \textit{RTSGAN} with the above modifications. More implementation details can be found in the Appendix \ref{appendix:incomplete}.

\textbf{Quantitative evaluation}.
We evaluate the utility of the synthetic data under the TSTR setting. For both PhysioNet and MIMIC-III, the downstream task is to predict in-hospital mortality.  
To be more comprehensive, we train different downstream classification models: 1)  \textbf{zeroRNN}, which imputes the missing values with zero and takes the concatenation of static features, dynamic features and missing patterns as the input features for an RNN classifier; 2) \textbf{lastRNN}, which imputes the missing values with the last valid observations (zero if no such observation exists) and takes the concatenation of static features, dynamic features and the time lags as the input features for an RNN classifier. Both two methods are implemented with a 2-layer GRU and imputation is performed after data scaling. Based on our observations that different data scaling methods on the synthetic data can affect the post-hoc classification performance, we apply both \textbf{min-max scaling} and \textbf{standard scaling} from scikit-learn\footnote{\url{https://scikit-learn.org}} in the experiments. 
In total, we have four ($2 \times 2 $) classification settings. 

The MIMIC-III benchmark has two strong baselines, so we adopt them to evaluate the utility of the synthetic datasets:
\begin{itemize}
\item[--]\textbf{Logistic Regression (LR)} uses a more elaborate version of the hand-engineered features described in \cite{lipton2015learning} to train a logistic regression model. We additionally tried min-max and standard scaling to normalize the features.

\item[--]\textbf{discreteLSTM} first re-samples the time series into regularly spaced intervals. Then it uses a 2-layer LSTM which takes new dynamic features and missing patterns for classification. This method uses standard scaling for numeric inputs.
\end{itemize}

In summary, we test four classification models on PhysioNet and seven models on MIMIC-III. The detailed descriptions can be found in the Appendix \ref{appendix:classification}. The area under the ROC curve (AUC) score is used to measure the classification performance because of the non-balanced datasets. We evaluate the utility of the synthetic datasets by averaging the classification performance over all models.

\begin{figure*}[t!]
  \centering
  \includegraphics[width=\linewidth]{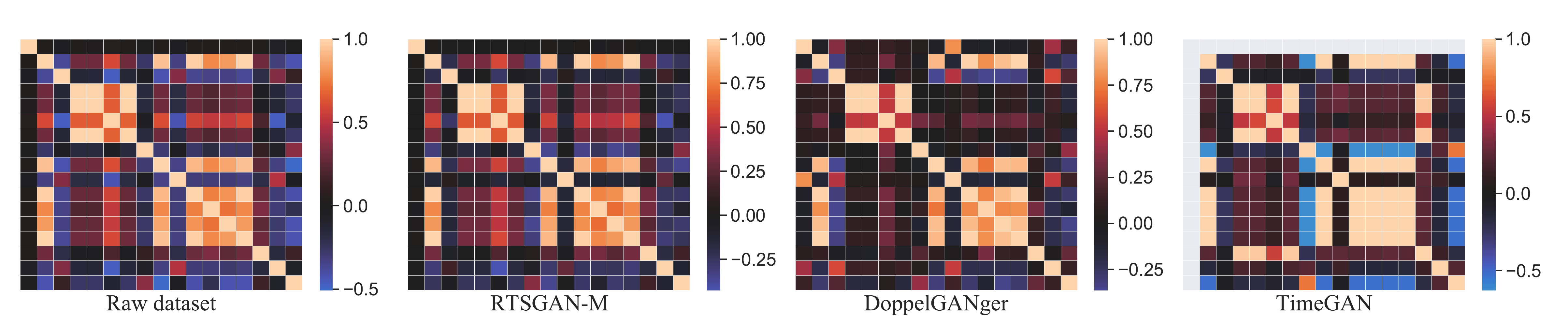}
  \caption{
  Pearson correlation heatmaps of missing rates of each pair of dynamic features on the MIMIC-III dataset and three  synthetic datasets.}\label{fig:heatmap_mimic}
\end{figure*}

\textbf{Qualitative evaluation}.
We also evaluate how well the synthetic data maintain the original missing patterns by visualization. We use the following two ways to visualize them.

a) {\bf 2D visualizations}. We sample from both the original and the synthetic datasets and calculate the missing rates of the dynamic features in each instance. Then we get the statistic vector of missing values for each instance and apply t-SNE and PCA for visualization. This can qualitatively assess the diversity of the synthetic missing patterns, and how close the distribution of the synthetic missing patterns is to that of the original missing patterns in 2D space.

b) {\bf Heatmap of the Pearson correlations}. Missing rates of the dynamic features are often correlated with each other. Some features are often measured together while some measures would be unlikely to happen at the same time. So we calculate the Pearson correlations about missing rates of each pair of dynamic features and visualize them by a heatmap. This helps us to understand the global properties of missing values in the dataset.

\subsubsection{Quantitative Results}

\begin{figure}[t!]
  \centering
  \includegraphics[width=1\linewidth]{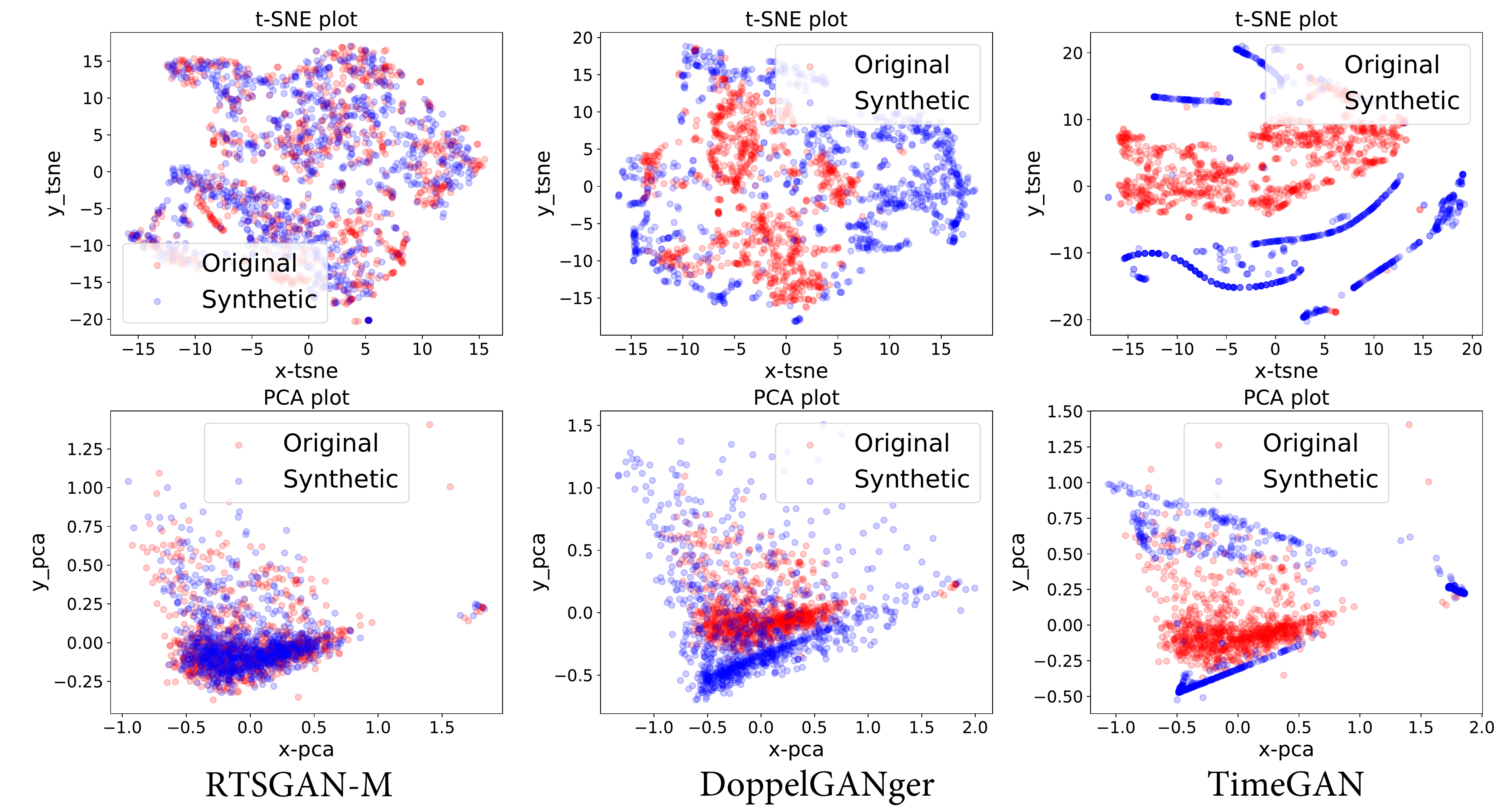}
  \caption{t-SNE (top) and PCA (bottom) visualization results on MIMIC-III dataset. Red and blue points denote the missing patterns of the original and synthetic instances respectively. }\label{fig:visual_mimic}
\end{figure}

Table~\ref{tab:mimic3} and Table~\ref{tab:compare2012} present the dataset utility comparison results on MIMIC-III and PhysioNet, respectively. We have the following observations:

a) In both Table~\ref{tab:mimic3} and Table~\ref{tab:compare2012}, we can see that \textit{RTSGAN} outperforms TimeGAN and DoppelGANger consistently when using different downstream classification methods. Although DoppelGANger introduces some targeted designs to address the challenges in time series data generation, it is still difficult to train a traditional GAN framework directly due to the high missing rates in the feature space of time series data. TimeGAN also combines the AE and GAN framework, but it does not perform well maybe because it is hard to optimize the generation of synthetic sequences of latent vectors especially when the lengths of time series data are long and variable. By contrast, the synthetic data generated by \textit{RTSGAN} and \textit{RTSGAN-M} have better utility for the mortality prediction task with the help of the encoder-decoder module in our design. 

b) \textit{RTSGAN-M} further outperforms \textit{RTSGAN}, especially for classification methods using standard scaling, which demonstrates that the proposed observation embedding and decide-and-generate decoder can better deal with the missing values. This also implies that simply treating missing patterns as additional binary features (used in \textit{RTSGAN}, TimeGAN and DoppelGANger) is not enough to generate high-utility time series with missing values.

c) We observed an interesting phenomenon in Table~\ref{tab:mimic3} that using standard scaling to normalize features generally leads to better AUC scores when the post-hoc models are trained on the original dataset. However, when they are trained on the synthetic dataset, using standard scaling will lead to a performance drop compared to using min-max scaling. 
Similar performance gap between using min-max scaling and standard scaling is also worth noting in Table~\ref{tab:compare2012} when training a \textit{lastRNN} on the synthetic dataset.
This phenomenon should be explored in the future work.

\begin{table}[]
\centering
\caption{Ablation study on the PhysioNet dataset. }
\label{tab:ablation2012}
\resizebox{\linewidth}{!}{\large
\begin{tabular}{cccccc}
\toprule
\multirow{2}{*}{Methods} & \multicolumn{2}{c}{min-max scaling} & \multicolumn{2}{c}{standard scaling} & \multirow{2}{*}{Avg.} \\
\cmidrule(lr){2-3} \cmidrule(lr){4-5}
                         & zeroRNN          & lastRNN          & zeroRNN           & lastRNN          &                      \\
\midrule
RTSGAN-M                 & 0.8152 & 0.8242 & 0.8134 & 0.7857 & 0.8096  \\
\midrule
w/o embedding  & 0.8040 & 0.8238 & 0.8067 & 0.7709 & 0.8013               \\
w/o two-step  & 0.8016 & 0.8213 & 0.7668 & 0.7076 & 0.7743 \\
\midrule
RTSGAN                 & 0.7866 & 0.8172 & 0.7371 & 0.6555 & 0.7491   \\
\bottomrule
\end{tabular}
}
\end{table}

\subsubsection{Qualitative Results}
Here, we examine the visualization results on MIMIC-III and do not show the visualization results on PhysioNet due to the same observations and the space limit.

Figure~\ref{fig:visual_mimic} shows the t-SNE and PCA visualization results on MIMIC-III. We can see that the missing pattern distribution of \textit{RTSGAN-M} is significantly closer to the original than those of TimeGAN and DoppelGANger. This indicates that \textit{RTSGAN-M} can better capture the missing data distribution than TimeGAN and DoppelGANger. In addition, the closer missing pattern distribution can also help to explain the better utility of synthetic data generated by our method.

Figure~\ref{fig:heatmap_mimic} shows the Pearson correlation heatmaps on MIMIC-III. From the heatmap of the original dataset, we can observe various correlation patterns between the missing rates of each pair of dynamic features. For instance, the missing rates of the 4th, 5th, 6th and 7th features are positively correlated because they are all measurements about ``Glascow'' coma estimations; The missing rate of ``Fraction inspired oxygen''  (the 3rd feature) and the missing rate of  ``Glascow coma scale total'' (the 6th feature) are negatively correlated because the fraction of inspired oxygen tends to be constant when continuously observing the patient's comatose state.

Compared to TimeGAN and DoppelGANger, the heatmap of the synthetic data from \textit{RTSGAN-M} looks the most similar to that of original data. It further confirms the superior performance of our proposed method in the incomplete time series data generation setting.

\begin{table*}[tbh!]
\centering
\caption{Ablation study on the MIMIC-III dataset. }
\label{tab:ablation_mimic}
\begin{tabular}{ccccccccc}
\toprule
\multirow{2}{*}{Methods} & \multicolumn{3}{c}{min-max scaling} & \multicolumn{4}{c}{standard scaling}     & \multirow{2}{*}{Avg.} \\
\cmidrule(lr){2-4} \cmidrule(lr){5-8}
                         & LR        & zeroRNN    & lastRNN    & LR     & zeroRNN & lastRNN & discreteLSTM &                      \\
\midrule
RTSGAN-M                 & 0.8209 & 0.8108 & 0.8068 & 0.7769 & 0.7753 & 0.7893 & 0.8221 & 0.8003 \\
\midrule
w/o time                 & 0.8270    & 0.8062     & 0.8032     & 0.7746 & 0.7475  & 0.7645  & 0.8142      & 0.7910               \\
w/o two-step             & 0.7982    & 0.7966     & 0.7855     & 0.7189 & 0.7218  & 0.7382  & 0.7928      & 0.7646               \\
\midrule
RTSGAN                   & 0.8009    & 0.7755     & 0.7682     & 0.6609 & 0.7450  & 0.7532  & 0.7940      & 0.7568  \\       
\bottomrule
\end{tabular}
\end{table*}

\subsection{Ablation Study}
To validate the effectiveness of each design in \textit{RTSGAN-M}, we conduct the ablation study on both the MIMIC-III and the PhysioNet dataset. We adopt the following variants of \textit{RTSGAN-M}: 1) ``w/o embedding'', in which we directly feed the inputs of the embedding module without using the observation embedding; 2) ``w/o two-step'', in which we use the decoder structure in \textit{RTSGAN} which outputs the missing patterns and values simultaneously instead of using the decide-and-generate decoder. As shown in Table~\ref{tab:ablation2012} and Table~\ref{tab:ablation_mimic},  both two designs can contribute to the utility of synthetic datasets. Relatively, the decide-and-generate decoder plays a more important role in \textit{RTSGAN-M} which better utilizes the informative missing patterns in data generation, without which the performance of \textit{RTSGAN-M} will drop significantly.

\section{Conclusions}
We proposed a novel time series generation framework \textit{RTSGAN} which produces realistic time series data with high utility for down-streaming tasks.
Besides generating time series data with variable lengths, to the best of our knowledge, this is the first work to explore generating time series with missing values. 
Moreover, equipped with the observation embedding and decide-and-generate decoder, the extended version \textit{RTSGAN-M} can effectively handle the missing values and generate high-utility synthetic data. The experiments on both complete and incomplete real-word time series datasets have shown the superiority of the proposed methods.

\bibliographystyle{IEEEtran}
\bibliography{main}

\appendix

\setcounter{totalnumber}{1}
\section{Reproducibility}

\subsection{Complete Time Series Generation}
\label{appendix:complete}
We use uniform sampling to train the autoencoder for 1000 epochs by Adam optimizer \cite{kingma2014adam} with learning rate 0.001 and $\beta_1 = 0.9, \beta_2 = 0.999 $. For the generation module, the dimension of the random vector $z$ just equals to the dimension of the latent space $(N+1)d_\mathrm{AE}$. We set the gradient penalty term $\lambda = 10$ for the WGAN as \cite{gulrajani2017improved} and train both the generator and the discriminator by RMSProp optimizer with learning rate 0.0001. We train the generator for 15000 iterations and update the discriminator five times per generator iteration. Batch size is 512 for training.

\subsection{Incomplete Time Series Generation}
\label{appendix:incomplete}
We use 3-layer GRU with hidden dimension 128 for \textit{RTSGAN}, \textit{RTSGAN-M} and baseline models on both PhysioNet and MIMIC-III. For \textit{RTSGAN-M}, We set $N_\mathrm{dec} = 2$ and the last hidden states of the encoder $\mathbf{h}_{l}^n$ would be fed into a fully connected layer followed by $\mathrm{LeakyReLU}$ activation before serving as the initial hidden states of the decoder $\hat{\mathbf{h}}_{0}^n$. We train the autoencoder for 800 epochs with batch size 128 using the teacher-forcing strategy. The threshold of whether each feature is missing is set by the overall missing rate on the training set. The rest part of model structures and training process are similar to \textit{RTSGAN} for complete time series generation. 

For the baseline models, we try different hyper-parameters according to what are reported in their papers \cite{yoon2019time, lin2020using}. 

\subsection{Downstream Classification Models for Incomplete Time Series}
\label{appendix:classification}
All downstream classification models are trained with the same hyper-parameters on both real and synthetic datasets. For PhysioNet, we use 2-layer GRU with hidden dimension 32 for both \textit{zeroRNN} and \textit{lastRNN} and use the final hidden state for classification. For MIMIC-III, we use 2-layer GRU with hidden dimension 16. We train these models for 40 epochs on the PhysioNet dataset, 30 epochs on the MIMIC-III dataset by Adam optimizer with learning rate 0.001.

\textit{LR} method on MIMIC-III computes six statistic features on seven different subsequences for each dynamic feature and uses these hand-crafted features to train a logistic regression model. 
The statistic features it uses include minimum, maximum, mean, standard deviation, skew and number of measurements. The seven subsequences include the full time series, the first 10\% of time, first 25\% of time, first 50\% of time, last 50\% of time, last 25\% of time, last 10\% of time. 
When a subsequence does not contain a measurement, the statistic features corresponding to that subsequence are later replaced with mean values computed on the training set.  L2 regularization with coefficient $C = 0.001$ is used when performing logistic regression. 

\textit{discreteLSTM} first re-samples the time series into regularly spaced intervals so that the time series become fixed-length. If there are multiple measurements of the same dynamic feature in the same interval, the method uses the value of the last measurement if it exists and a pre-specified ``normal'' value otherwise. After imputation, standard scaling is applied to the continuous features. The classification model is a 2-layer LSTM with hidden dimension 16 and trained for 30 epochs. The implementations of \textit{LR} and \textit{discreteLSTM} can be found in the git repository\footnote{https://github.com/YerevaNN/mimic3-benchmarks} of the MIMIC-III benchmark \cite{harutyunyan2019multitask}.

\end{document}